\documentclass[11pt,a4paper]{article}
\usepackage[hyperref]{acl2020}
\usepackage{times}
\usepackage{latexsym}
\usepackage{csquotes}
\usepackage{graphicx}
\usepackage{xspace}
\usepackage{subcaption}
\usepackage{graphicx}  
\usepackage{enumitem}

\usepackage{microtype}
\usepackage{soul}
\usepackage{booktabs}
\aclfinalcopy 

\newcommand\ie{i.\,e.\xspace}
\newcommand\eg{e.\,g.\xspace}

\title{Practical Annotation Strategies for Question Answering Datasets}

\author{Bernhard Kratzwald$^\spadesuit$ \quad Xiang Yue$^\diamondsuit$ \quad Huan Sun$^\diamondsuit$ \quad Stefan Feuerriegel$^\spadesuit$ \\
        $^\spadesuit$ Chair of Management Information Systems, ETH Zurich \\
        $^\diamondsuit$  Department of Computer Science and Engineering, The Ohio State University \\
        {\tt \{bkratzwald, sfeuerriegel\}@ethz.ch \{yue.149, sun.397\}@osu.edu} \\
}

\date{}

\begin{document}
\maketitle
\begin{abstract}
Annotating datasets for question answering (QA) tasks is very costly, as it requires intensive manual labor and often domain-specific knowledge. Yet strategies for annotating QA datasets in a cost-effective manner are scarce. To provide a remedy for practitioners, our objective is to develop heuristic rules for annotating a subset of questions, so that the annotation cost is reduced while maintaining both in- and out-of-domain performance. For this, we conduct a large-scale analysis in order to derive practical recommendations. First, we demonstrate experimentally that more training samples contribute often only to a higher in-domain test-set performance, but do not help the model in generalizing to unseen datasets. Second, we develop a model-guided annotation strategy: it makes recommendation with regard which subset of samples should be annotated. Its effectiveness is demonstrated in a case study based on a domain customization of QA to a clinical setting. Here, remarkably, annotating a stratified subset with only 1.2\,\% of the original training set achieves 97.7\,\% of the performance as if the complete dataset was annotated. Hence, the labeling effort can be reduced immensely. Altogether, our work fulfills a demand in practice when labeling budgets are limited and where thus recommendations are needed for annotating QA datasets more cost-effectively.

\end{abstract}

\section{Introduction}

State-of-the-art question answering~(QA) over content repositories is commonly based on machine comprehension \cite[\eg][]{Chen.2017,Seo.2017,yu2018qanet,Kratzwald.2019a}: here a neural network extracts an answer to a question from a given context document. The use of neural networks in QA has widespread implications. Foremost, learning such models requires a large amount of training data, yet data availability in practice is limited (\ie, available datasets stem primarily from open-domain settings in English). 
Annotating new datasets is an extremely costly undertaking that requires intensive manual labor and expert knowledge \cite{Molla.2007,pampari2018emrqa}. Owing to this, the labeling budgets that are available to practitioners often put a barrier to annotating large-scale QA datasets, and yet strategies for annotating QA datasets in a cost-effective manner are scarce.  


Given the above reasons, our objective is to guide practitioners in annotating QA datasets more cost-effectively. For this, we aim at developing a set of heuristics that suggest a subset of questions subject to annotation. The latter involves a manual and thus costly step in which annotators determine the correct answer for a given questions. This gives rise to a trade-off: fewer annotations reduce the overall labeling cost, yet a larger number of annotations should potentially facilitate learning. By choosing a stratified subset, we hope that our annotations are more cost-effective. Here we adhere to constraints from practice: (i)~annotating answers is costly, whereas questions come from users and thus at low cost; and (ii)~the annotation strategy should be used a~priori, that is, before deploying the system and before any ground-truth labels are available. 


We derive practical recommendations for the above task by performing a suite of large-scale experiments (amounting to more than 300 days of computational time). These unravel determinants of performance in neural QA (\ie, BERT) and, based on them, we design practical guidelines. Specifically, we proceed as follows. In Sec.~\ref{sec:size}, we investigate how the \emph{dataset size}  contributes to both the in-domain and out-of-domain performance (\ie, the latter refers to models trained on dataset $A$ but that are then evaluated on unseen datasets $B,C,\ldots$). In Sec.~\ref{sec:quality}, we develop different \emph{stratifications} via a model-guided annotation strategy. 

In Sec.~\ref{sec:context}, we examine the influence of \emph{context diversity}, \ie, how important it is to annotate a diverse set of context documents. Altogether, this yield heuristics that steer annotators towards a stratified subset of samples and thereby reduce the labeling effort while largely maintaining the original performance both in- and out-of-domain.  

Our main findings are summarized as follows:
\begin{itemize}[topsep=0pt,itemsep=-1ex,partopsep=1ex,parsep=1ex,leftmargin=2.5ex]
    \item Surprisingly, when increasing the size of a dataset, we find that neural QA stops generalizing to unseen datasets before they stop improving on the dataset used during training.\footnote{For this purpose, we define \textquote{saturation} as reaching 99.5\,\% of the performance of the model that was trained on the full dataset: When training and evaluating on different 90 to 10 split-ratios of the SQuAD training set we saw a standard deviation of approximately 0.5\%, we thus concluded that reaching 99.5\% of the full models performance indicates a saturation and the remaining 0.5\% are up to random fluctuations in the training process.} This implies that the dataset size could be significantly reduced without negatively affecting their generalization power to out-of-domain datasets.
    \item We show that datasets following our model guided annotation strategies achieve the same in-domain and out-of-domain performance, with only 65\,\% of the original training samples.
    \item We demonstrate the effectiveness of our findings in a case study over a domain-specific clinical QA dataset (Sec.~\ref{sec:casestudy}). In our numerical experiments, conventional annotations are constantly outperformed by our proposed strategies. Interestingly, we are able to reach 97.7\,\% of the performance from the complete dataset -- even though we annotate a mere 1.2\,\% of the original samples. 
\end{itemize}

\begin{figure*}[h]
    \centering
    \includegraphics[width=0.9\textwidth]{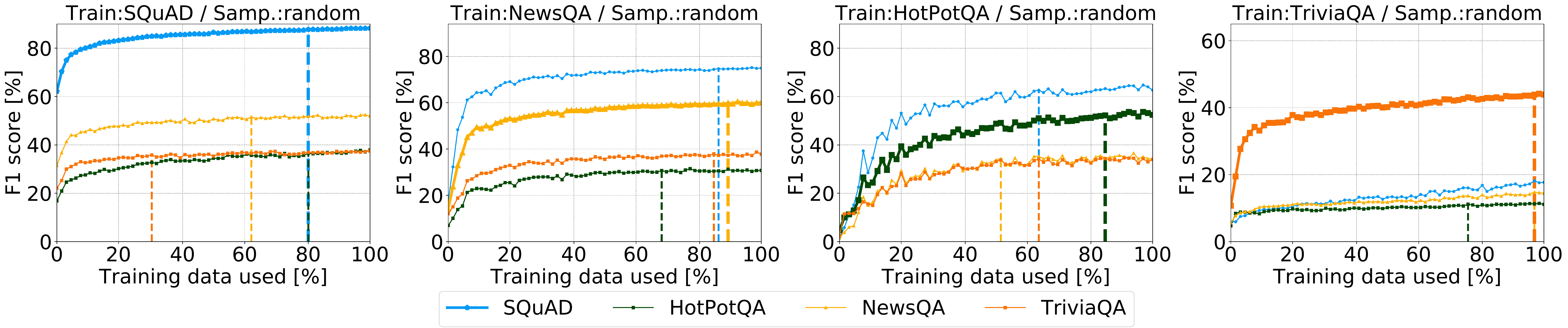}
    \caption{Influence of the training data size on performance and generalization. The bold line indicates the performance on the test-set of the dataset we trained on, the three non-bold lines indicate the performance on datasets not seen during training (generalization). The vertical bars indicate where training saturated: We define saturation as reaching $99.5\,\%$ of the performance of the model that was trained on the full dataset.}
    \label{fig:main}
\end{figure*}
\begin{figure*}[h]
    \centering
    \includegraphics[width=0.9\textwidth]{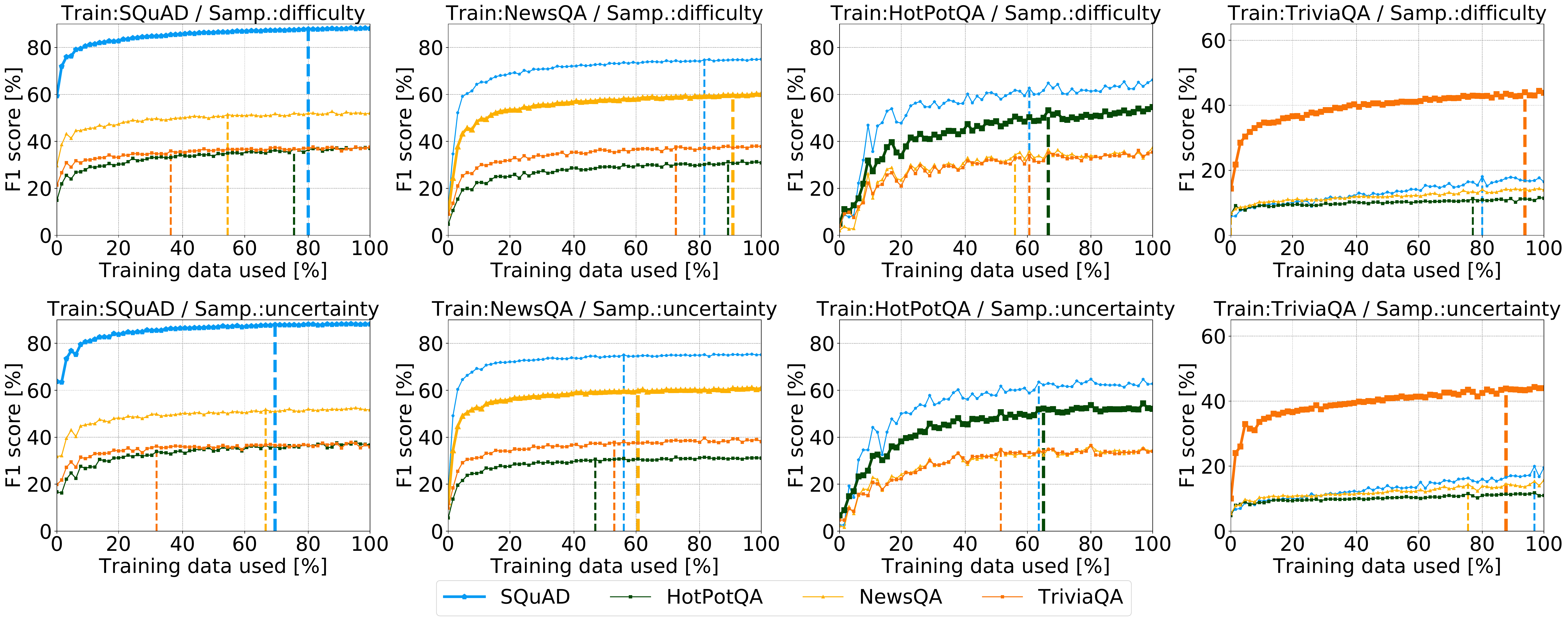}
    \caption{Influence of the training data size on performance and generalization. Here samples were not chosen randomly but following model-guided annotation strategies.}
    \label{fig:quality}
\end{figure*}

\section{Related Work}
Earlier research on question answering datasets focused on the response to adversarial and thus manipulated samples \cite{Jia.2017,Mudrakarta.2018}. \Citet{Kaushik.2018} examined the relative importance of the context, i.e., how many questions can be answered directly without it. \citet{Sugawara.2018} counted the ratio of questions within benchmarks classified as hard vs. easy (based on similarity and entity type heuristics). Other research \cite{talmor-berant-2019-multiqa,fisch-etal-2019-mrqa} studied the generalization power of QA datasets, \ie, how models trained on one dataset generalize to other unseen datasets. However, prior works have ignored the question of how the in-domain and out-of-domain performance is influenced by the dataset size. Owing to this, it is also unclear which subset of data samples should be annotated (to reduce the labeling effort while maintaining performance). 

The approach proposed in this paper is rooted in active learning, assuming an unlabeled pool of resources for which we can request labels in every iteration. It is different from curriculum learning~\cite{bengio2009curriculum} in which data from a labeled dataset is subject to non-uniform sampling, whereas our task is to infer a subset from an unlabeled pool of samples.

\section{Experimental Setting}
We ran an extensive set of experiments with 7 GPUs corresponding to a total computation time of approximately 300 days. The source code to reproduce our results is attached to this submission. 

\textbf{Datasets:} For our analysis we draw upon four common QA datasets: (i) SQuAD \cite{rajpurkar2016squad}, (ii) NewsQA \cite{trischler-etal-2017-newsqa}, (iii) HotPotQA \cite{yang-etal-2018-hotpotqa} (iv) TriviaQA \cite{joshi-etal-2017-triviaqa}. For details, we refer to Appendix~A.

\textbf{Model:} We use BERT \cite{devlin2019bert} throughout our analyses. In our domain-specific case study, we demonstrate that our approach also transfers to other types of models and extend our analysis to the DocReader \cite{Chen.2017}. 

\textbf{Experimental design:} In our experiments, we investigate how the number of training samples influences the performance on the dataset we choose for training $\mathcal{D}_P$ and the generalization to datasets unseen in training $\mathcal{D}_G$. Therefore, all experiments follow four steps: (i) Initially, we train a BERT model on a random sample of $b=1.5\,\%$ of the training-set of dataset $\mathcal{D}_P$. (ii)~After training, we evaluate the \emph{performance} on a hold-out fraction of dataset $\mathcal{D}_P$. (iii) To evaluate the \emph{generalization} power we also evaluate its performance on test-sets of datasets that we did not use in training, \ie, $\mathcal{D}_G$, (iv) We increase the number of training samples by $b$. We then repeat steps (ii) to (iv) until all of the training data of $\mathcal{D}_P$ is used. 

We reiterate that $\mathcal{D}_P$ measures the performance on questions from the training population, whereas $\mathcal{D}_G$ measures the performance on unseen datasets and thus generalization. The experiments described in the next three sections differ in how we select new samples in step~(iv), \eg, randomly vs. by context diversity. We averaged all results across three separate runs. 

\section{How does the Size of a Dataset Impact Performance and Generalization?}
\label{sec:size}

Now we investigate how the performance is influenced by the number samples used during training. For this, our above experimental setting returns the performance as a function of the training-set size. This is done for all comparisons between the datasets; see Fig.~\ref{fig:main}. The vertical bars indicate where training saturated. Here we define \emph{\textquote{saturation}} as reaching $99.5\,\%$\ of the performance of the model that was trained on the full dataset. Simply put, it hints when the model stops learning. 

\textbf{Results:} (1)~\underline{Performance on $\mathcal{D}_P$}: For SQuAD, NewsQA, and HotPotQA, our results show that around 85 to 90 percent of training samples are sufficient for the model in order to reach saturation on their own test set. Adding more samples than that barely affects the performance. (2)~\underline{Generalization to $\mathcal{D}_G$}: In comparison to before, performance saturates on datasets not seen in training (\ie, generalization power) usually earlier. Sometimes models stop generalizing to unseen datasets after using only 30\,\% of the training data. (3)~For  \underline{TriviaQA}: saturation on the training set and generalization to unseen sets are both reached very late. This is likely to be caused by the fact that TriviaQA is the only dataset that was annotated via distant supervision, whereas all other datasets have been manually annotated by humans. Furthermore, this dataset generalizes poorly to others (cf. final performance of the three other datasets when trained on TriviaQA). 

\textbf{Recommendations:} The neural QA model stops generalizing to unseen datasets before it saturates on the dataset used during training. Hence, larger datasets are barely helpful. To this end, the gap between generalization and saturation gives strong evidence that current open-domain QA datasets could be significantly reduced in size without affecting their performance. We find that datasets with a higher annotation quality (manual vs. distantly supervised annotation) both saturate faster and generalize better. Hence, practitioners could save a labeling effort without facing downturns in generalizability, by \emph{first annotating a test sample and then stop data annotation once saturation on that test-set is observed.}

\section{How Does a Stratified Annotation Help Learning?}
\label{sec:quality}

Previously, the selection of questions for annotation was random, while we now experiment with stratified approaches from active learning.\footnote{In every iteration (iv) of our experiment, we train a model on the currently labeled data and use that model to score the remaining samples with our annotation strategies. We then select those samples with the highest score.} Thereby, we test if stratified annotators can reach saturation earlier by showing them an estimate of how helpful their annotations are. Therefore, we design two model-guided annotation strategies. 

\textsc{Question Difficulty:} The first strategy is based on the idea that simple questions can be answered without learning proper semantics~\cite{Sugawara.2018}. An example would be to ask a question about \emph{who} did something when there is only one entity mentioned in the context document. A model would not require to learn the semantics of the full question. We detect such questions by predicting the answer for every sample twice: once using the complete question $q$, and once using only the first three words of $q$. If the predicted answers of both are equal, the question is labeled as {easy} and otherwise as {hard}. We then sample those questions labeled as hard first, and only subsequently sample questions labeled as easy.

\textsc{Model Uncertainty} is captured by Shannon's entropy. In detail, we predict the answer span for a question within the context document and average the entropy of the start and end prediction for the answer. In every iteration, we select the samples with the highest model uncertainty. 

\textbf{Results:} (1)~\underline{Question difficulty} sampling: Results are shown in the top row of Fig.~\ref{fig:quality}. When comparing it with the random sampling approach in Fig.~\ref{fig:main}, we can see that saturation on the training dataset is improved. On average, saturation is reached with using 4.9\,\% fewer samples and generalization with 9.5\,\% less. (2)~\underline{Model uncertainty} sampling: This positively affects both saturation and generalization. In detail, we need around 16.9\,\% less data to saturate on the training dataset and around 10.2\,\% less data to saturate on other datasets (generalization) than by random sampling. For TriviaQA, the results are worse which is a result from the fact that ground-truth annotations are known to fairly noisy (cf. discussion earlier). 

\textbf{Recommendations:} The above annotation strategies reduce the gap between the points where saturation on the training dataset and on other datasets is reached. In practice, such strategies are easy-to-compute and, by being model-guided, are ensured to be flexible. Furthermore, they could be displayed to annotators on-the-fly to provide them with an estimate of their annotation quality. \emph{Hence, practitioners should employ model-guided annotation strategies (especially uncertainty sampling) as these help in reaching the same performance level on both the original datasets and unseen dataset with considerably fewer samples. }

\section{What is the Benefit of Annotating Diverse Contexts First?}
\label{sec:context}

\begin{figure}[h]
    \centering
    \includegraphics[width=0.48\textwidth]{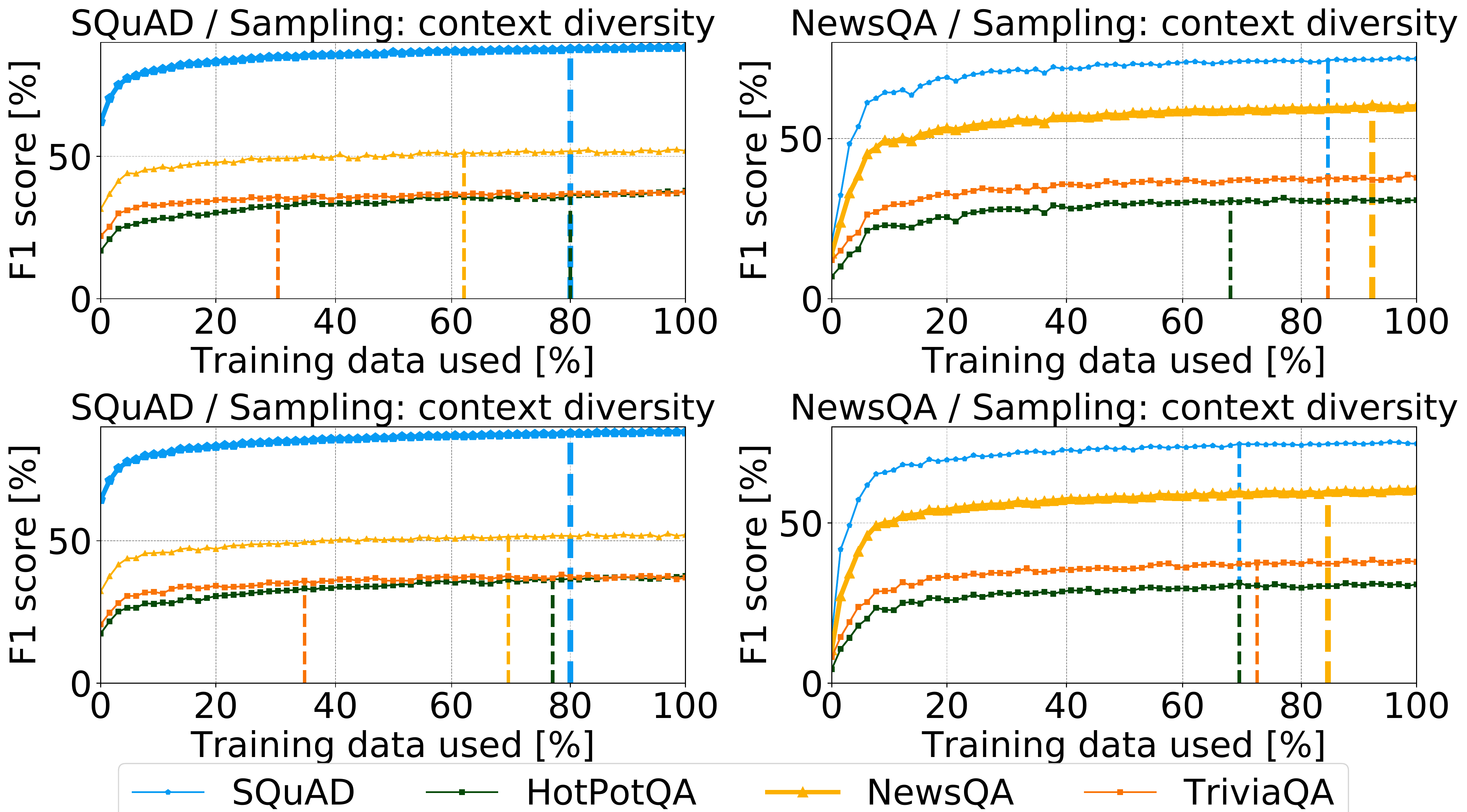}
    \caption{Comparison of random sampling (top row) against context-document sampling (bottom row).}
    \label{fig:contextdiv}
\end{figure}

As a robustness check, we evaluate whether there is a benefit from annotating a diverse set of context documents. This analysis is based on SQuAD and NewsQA; the reason is that these datasets contain multiple question-answer annotations for a single context document. This allows us then to compare two sampling strategies: (i)~random sampling and (ii)~context-document sampling whether a round-robin approach is used to first sample all context documents. 

\textbf{Results:} The results are shown in Fig.~\ref{fig:contextdiv} of our appendix. For NewsQA, sampling questions with more diverse contexts benefits both saturation and generalization. However, the performance improvements are not as substantial as for the model-guided annotation strategies. For SQuAD, we barely observe an improvement. Both results can be explained by the nature of the datasets: NewsQA consists of annotated newspaper articles that cover a wide spectrum of different topics and author styles, whereas context documents in SQuAD are given by paragraphs that are primarily extracted from a small set of Wikipedia articles. 

\textbf{Recommendations:} \emph{For corpora with diverse documents, annotations should be divided upon a divers set of context documents.} This might prevail the classical approach of annotating multiple question-answer pairs for a single document. This is further substantiated in our case study.

\section{Case Study: Clinical QA}
\label{sec:casestudy}


The value of our above recommendations is demonstrated based on a domain-specific setting. For this purpose, we draw upon clinical QA, specifically the emrQA dataset \cite{pampari2018emrqa}. It provides a large-scale QA dataset (around 1 million QA pairs) with clinical notes from electronic medical records. Our implementation is based on DocReader \cite{Chen.2017}.\footnote{We also experimented with BERT but it performance was inferior in our case study and was thus omitted for brevity.} For a detailed description we refer to our Appendix.

We operationalize the above recommendations (\ie, annotate diverse context documents and select high-quality questions) via the following annotation strategies: (1)~\textsc{conventional sampling}: we annotate 50 questions per context for 148 documents. (2)~\textsc{context document sampling}: we annotates 25 questions per context, but use 296 distinct documents; (3)~\textsc{Model uncertainty} we use entropy to annotate those 25 questions per context that have the highest model uncertainty. All strategies result in roughly 7,500 annotated samples (approx. 1.2\,\% of the original training data). For comparison, we list (4)~\textsc{full dataset} which refers to the performance when annotating the complete training data of $\sim$ 620,000 QA pairs. 

The corresponding performance is listed in Table.~\ref{tbl:emrQA}. \textsc{context document sampling} strategy outperforms \textsc{conventional sampling}, thereby pointing towards benefits from annotating more diverse set of context documents.\footnote{The relatively high performance of the conventional sampling approach is likely caused by the fact that the emrQA dataset was generated from question templates. It may contain many duplicated questions with the similar format and semantic meaning.} The performance is further improved when using a model-based annotation strategy (\ie, \textsc{model uncertainty}). In sum, our annotation strategies yield compact datasets yet that achieve a remarkable performance: in fact, 1.2\,\% of the data are sufficient to reach up to 97.7\% of the performance level when annotating the complete dataset.

\begin{table}[t]
\footnotesize
\centering
\resizebox{\linewidth}{!}{%
\begin{tabular}{lcc}
\toprule
\textbf{Annotation strategy} & \textbf{Dev} & \textbf{Test} \\ 
\midrule
Conventional sampling & 81.39/90.61 & 81.59/90.94  \\ 
Context-document sampling & 83.14/92.19 & 83.56/92.73  \\ 
\ + model uncertainty sampling & 83.22/92.26 & 83.77/92.45   \\
\midrule
Full dataset & 86.43/94.44 & 86.94/94.85 \\ \bottomrule
\end{tabular}%
}
\caption{Performance (stated: Exact Match/F1) that is achieved when applying our annotation strategies to a domain-specific use case (here: clinical QA).}
\label{tbl:emrQA}
\end{table}

\bibliography{arxiv}
\bibliographystyle{acl_natbib}

\appendix
\section{Datasets}
 For our analysis we draw upon the following QA datasets: 
 \begin{enumerate}
     \item SQuAD \cite{rajpurkar2016squad}: The Stanford Question Answering Dataset contains over 100,000 samples. Question-answer pairs are annotated on a small set of Wikipedia paragraphs.
     \item NewsQA \cite{trischler-etal-2017-newsqa}: Question-answer pairs are annotated for CNN news articles, that are longer than Wikipedia paragraphs.
     \item  HotPotQA \cite{yang-etal-2018-hotpotqa}: HotPotQA is a dataset requiring multi-hop reasoning over several paragraphs in order to answer a question. The dataset also provides additional paragraphs to distract the model and make the prediction harder. 
     \item  TriviaQA \cite{joshi-etal-2017-triviaqa}: A dataset composed of trivia questions with their answers. Multiple bing-snippets are provided as context documents. The position of the answer within then context document has been annotated by using distant supervision rather than human annotations.
     \item For our case study we use emrQA \cite{pampari2018emrqa}: A large-scale clinical QA dataset created by a generation framework from a small number of expert-annotated question-templates and existing clinical corpora annotations.  
 \end{enumerate}
 
 \subsection{Reprocessing for Experiments}
 
All datasets have been downloaded in the MultiQA format which includes a unanimous prepossessing for all datasets as described in \cite{talmor-berant-2019-multiqa}.\footnote{Available from \url{https://github.com/alontalmor/MultiQA} (Last opened: Dec. 7th, 2019)} We then converted the data to the BEART-readable SQuAD format.\footnote{Using this script: \url{https://github.com/alontalmor/MultiQA/blob/master/convert_multiqa_to_squad_format.py} (Last opened: Dec. 7th, 2019)} A bash-script automating this process is attached to this submission.

 \subsection{Reprocessing for Case-Study}
 EmrQA \cite{pampari2018emrqa} contains 5 subsets. In this paper, we just focus on the largest one: the \emph{Relation} dataset, which contains $\approx$1 million QA pairs after filtering the questions whose answer lengths are more than 30. We split the dataset into train, dev, test based on the contexts following the ratio of 7:1:2.

\section{Hyperparameter Configurations}
Please not that the source code to reproduce the results shown in our paper is attached to this submission.

We use the official BERT implementation\footnote{Available from \url{https://github.com/google-research/bert} (Last opened: Dec. 7th, 2019)} in tensorflow in our experiments. Hyper-parameters are used as reported in \cite{devlin2019bert} for SQuAD, where we only reduced the training batch size to 12 and the sequence length to 360 for memory issues.
\end{document}